\documentclass{bmvc2k}
\usepackage{graphicx}
\usepackage{amsmath}
\usepackage{amssymb}
\usepackage{pgf}
\usepackage{bbm}
\usepackage{multirow}
\usepackage{lipsum}
\usepackage{hhline}
\usepackage{subcaption}
\usepackage{caption}

\graphicspath{{images/}}

\title{Three for one and one for three:\\
Flow, Segmentation, and Surface Normals}

\addauthor{Hoang-An Le}{hoang-an.le@uva.nl}{1}
\addauthor{Anil S. Baslamisli}{a.s.baslamisli@uva.nl}{1}
\addauthor{Thomas Mensink}{thomas.mensink@uva.nl}{1}
\addauthor{Theo Gevers}{th.gevers@uva.nl}{1}

\addinstitution{
 Computer Vision Group, \\
 Informatics Institute, \\
 University of Amsterdam,\\
 the Netherlands
}

\runninghead{Le \etal}{Three for one and one for three}

\def\etal{\emph{et al}\bmvaOneDot}

\newcommand{\mysubsubsection}[1]{%
    \vspace{-3mm}
    \subsubsection*{#1}\vspace{-3mm}
}
\begin{document}

\maketitle

\begin{abstract}
Optical flow, semantic segmentation,  and surface normals represent different information modalities,
yet together they bring better cues for scene understanding problems. In this paper, we study the 
influence between the three modalities: how one impacts on the others and their efficiency in 
combination. We employ a modular approach using a convolutional refinement network which is trained
supervised but isolated from RGB images to enforce joint modality features. To assist the training process, we create a 
large-scale synthetic outdoor dataset that supports dense annotation of semantic segmentation,
optical flow, and surface normals. The experimental results
show positive influence among the three modalities, especially for objects' boundaries, 
region consistency, and scene structures. 
\end{abstract}

\section{Introduction}

    Optical flow, semantic segmentation, and surface normals represent different aspects
    of objects in a scene i.e. object motion, category, and geometry. 
    While they are often approached as single-task problems, their combinations
    are of importance for general scene understanding as humans also
    rarely perceive objects in a single modality. As different information sources provide 
    different cues to understand the world, they could also become complementary to each 
    other. For example, certain objects have specific motion patterns (flow and semantics), 
    an object's geometry provides specific cues about its category (surface normals and semantics),
    and object's boundary curves provide cues about motion boundaries (flow and surface normals).

\begin{figure}[t]
        \centering
        \includegraphics[width=.9\linewidth]{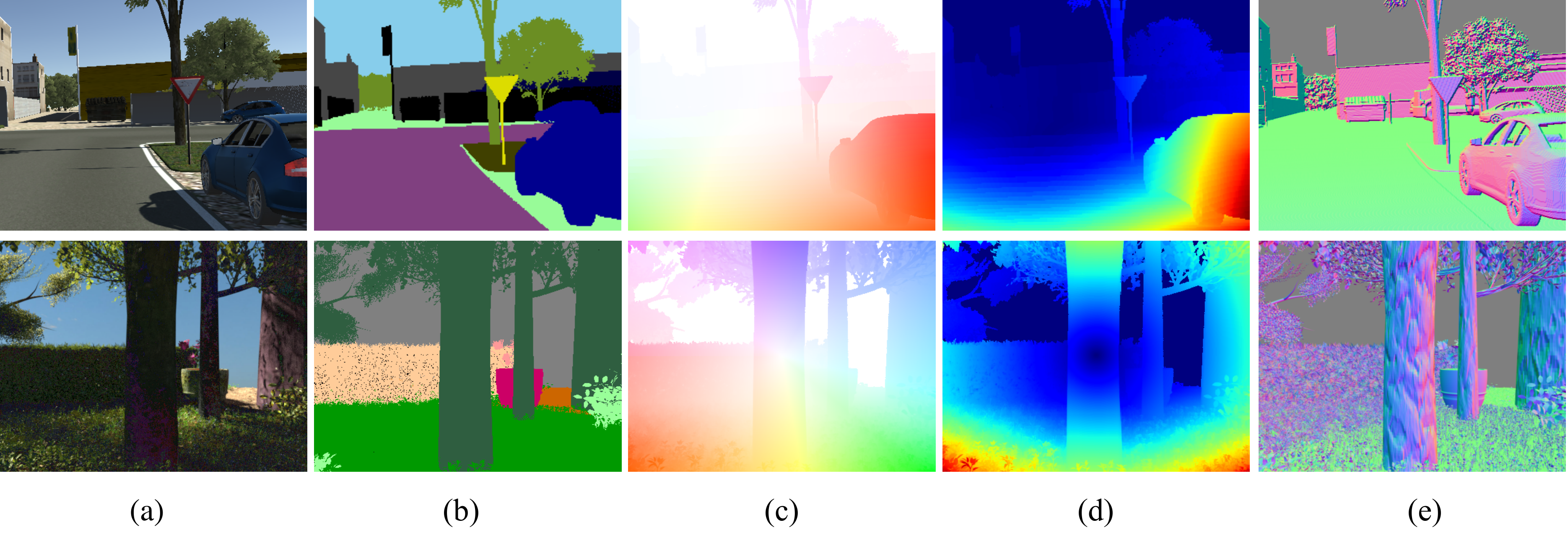}
        \vspace{1mm}
        \caption{Visualization from VKITTI \cite{virtualKITTI} (\emph{top}) and Nature (\emph{bottom}) dataset. 
        From  left to right: 
        (a) RGB image, 
        (b) semantic annotation, 
        (c) color-coded optical flow, 
        (d) corresponding flow magnitude;
        (e) surface normal.
        }
        \label{fig:gt_vis}
    \end{figure}

    Scene-based optical flow estimation is a challenging problem because of complicated
    scene variations such as texture-less regions, large displacements, strong 
    illumination changes, cast shadows, and specularities. As a result, optical flow
    estimation tends to perform poorly in homogeneous areas or around objects' boundaries. 
    Another hindrance for many optical flow estimators is the common assumption of spatial 
    homogeneity in the flow structure across an image~\cite{Sevilla-Lara_2016_CVPR}. 
    That assumption poses difficulties as different objects have different motion patterns:
    objects closer to the viewer have stronger optical flow;
    independent objects have their own flow fields, while static objects follow the camera's motion
    (Figure~\ref{fig:gt_vis}).
    Thus, by modeling optical flow based on image segmentation, one could improve flow accuracy, 
    especially at objects' boundaries~\cite{Bai2016,Sevilla-Lara_2016_CVPR}.
    
    The goal of image segmentation is to partition an image into different parts that share common properties.
    In particular, semantic segmentation assigns to each 
    image pixel the object category to which the pixels belong. It is a challenging task
    especially in videos, due to inherent video artifacts such as motion blur,
    frame-to-frame object-to-object occlusions and object deformations. 
    As optical flow encodes temporal-visual information of image sequences, it is
    often exploited to relate scene changes over time~\cite{Lee2016, fgfa_iccv17, zhu17dff}.
    Yet, optical flow, as an approximation to image motion~\cite{Beauchemin1995}, 
    also encodes the 3D structure of a viewed scene. If the camera's translation is 
    known beforehand, an optical flow image can be used to recover the scene depth~\cite{Baraldi1989}.
    Considering a moving camera, closer objects appear with stronger motion fields than 
    distant ones, independent moving objects have prominent motion patterns compared to the 
    background, and different object shapes generate different motion fields because of depth
    discontinuities. 
    Therefore, temporal and structural information provided by 
    optical flow can guide semantic segmentation by providing cues about 
    scene ambiguities. 
    
    Surface normals, on the other hand, represent changes of depth,
    i.e. the orientation of object surfaces in 3D space. Thus, they are independent of
    illumination effects or object textures. That information is particular helpful for 
    texture-less objects, regions of strong cast shadows, or scenes of less visibility. 
    Additionally, object boundaries provide cues about both motion and semantic boundaries.
    Therefore, surface normal information is expected to assist the optical flow estimation process
    by providing various cues.
    Figure~\ref{fig:gt_vis} illustrates the relationship of 
    object boundaries depicted in segmentation and 3D structure cues in optical 
    flow and surface normal images. 
    
    In this paper, we study the mutual interaction of optical flow, semantic segmentation, 
    and surface normals and analyze their contribution to each other.  
    We employ a modular approach and adapt a convolution based 
    supervised refinement network to examine the efficiency of joint features 
    from the different modalities.
    
    Large-scale datasets with optical flow and surface normal ground truths are hard to obtain. 
    Apart from semantic segmentation, it is not intuitive for humans to manually come up with
    pixel-wise annotations for optical flow and surface normals. Hence,
    only a small number of real-world datasets provide annotations for either optical flow 
    or surface normals, but not for both. Although one may construct surface normals from depth images,
    the process tends to produce unwanted artifacts. Thus, 
    synthetic data is favorable.
    To that end, we construct a large-scale synthetic dataset of nature scenes such as 
    gardens and parks, that provides ground-truth annotations for optical flow, 
    surface normals and semantic segmentation. 
    
    In summary, our contributions are: (1) the connection among the three modalities 
    (optical flow, semantic segmentation and surface normals), 
    (2) adapting a convolutional based supervised refinement network to improve one of 
    the three using the other two, 
    (3) an experimental study to estimate all three in a joint fashion,
    (4) a large-scale scene-level synthetic images of nature scenes such as gardens and parks 
    with  ground-truth annotations for optical flow, surface normals and semantic segmentation.
    

           
\section{Segmentation, Flow, and Surface Normals}
\label{sec:related_work}

    \subsection{Related Work}
    
    In this section, we review the work on optical flow, semantic segmentation,
    and surface normals and how they are mostly targeted as single tasks.

    \textbf{Optical flow} is defined as the apparent motion field
    resulted from an intensity displacement in a time-ordered sequence of images.
    It is an approximation to image motion, because estimating optical flow 
    is an ill-posed problem~\cite{Beauchemin1995}. To model the displacement 
    of image intensities that are caused solely by the objects' motion in 
    the physical world, several priors are derived to constrain the problem. Two of
    the most exploited ones are the assumptions of brightness constancy and 
    Lambertian surface  reflectance~\cite{Fortun2015, Wedel2011, Beauchemin1995}. 
    Besides, \cite{Black1996} makes use of
    robust statistics to promote discontinuity-preservation. Many popular methods also
    apply coarse-to-fine strategies \cite{EpicFlow,Xu2012,Brox_2011}. 
    On the other hand, deep convolutional neural networks (CNNs) are dominating the
    field more recently. For instance, \cite{Weinzaepfel_2013} applies a coarse-to-fine
    strategy with the help of a CNN framework. Then, Dosovitskiy~\etal~\cite{Dosovitskiy2016}
	proposes an end-to-end CNN called FlowNet, which is later improved by 
	Ilg~\etal~\cite{Ilg2016} to perform state-of-the-art optical flow estimations. 
	
	\textbf{Semantic Segmentation} is vital for robot vision and 
	scene understanding tasks as it provides pixel-wise annotations to scene properties.
	Traditional methods approach the problem by engineering hand-crafted features and perform 
	pixel-wise classification with the help of a classifier~\cite{Shotton2009,Csurka2011}.
	Other works try to group semantically similar pixels~\cite{Comaniciu2012,Mori_2004}.
	Like most of the computer vision tasks, semantic segmentation also benefits from powerful
	CNN models. 
	After the pioneering work of~\cite{Long2015}, many other deep learning 
	based methods are proposed such as~\cite{Yu2016, Chen2016}. 
	
	\textbf{Surface Normals} provide information about an object's 
	surface geometry. Traditional methods that infer 3D scene layout from single
	images rely on primitives detection such as oriented 3D surfaces~\cite{Hoiem2007}
	or volumetric primitives~\cite{Gupta2010}. Their performance depends on the
	discriminative appearance of the primitives~\cite{Fouhey2013}. In the context of deep learning,
	Wang~\etal~\cite{Wang2015} proposes a method to predict surface normals from a single color
	image by employing a scene understanding network architecture with physical constraints;
	Bansal~\etal~\cite{Bansal2016} predicts surface normals and use them as an intermediate
	representation for 3D volumetric objects for model retrieval. Eigen and 
	Fergus~\cite{Eigen2015} designs an architecture that can be used for predicting each of
	the three modalities, including depth, surface normals, and semantic segmentation, in a separating
	manner.
	
    \subsection{Cross-modality Influence}
    
    \mysubsubsection{Three for Optical Flow}
    
    Image motion, although varies across image regions, is often treated in the same way by
    many optical flow methods. Semantic segmentation provides a way to partition an image
    into different groups of predefined semantic classes. Hence it provides optical flow with
    information of object boundaries, and helps to enforce motion consistency within similar
    object regions. Similar ideas employed by \cite{Bai2016} with object 
    instance-segmentation or \cite{Sevilla-Lara_2016_CVPR} with 3 semantic classes 
    (things, planes, and stuff) have shown to improve optical flow accuracy.
    
    On the other hand, surface normals represent the orientation of objects' surfaces in 3D space.
    They contain geometry information that is invariant to scene lighting and objects' appearance,
    which is rendered useful for optical flow in case of intricate lighting such as cast shadows,
    texture-less regions, specularities, etc. Additionally, surface normals can be beneficial for 
    optical flow at depth order reasoning, and may improve accuracy at occlusion boundaries.
	
    \mysubsubsection{Three for Semantic Segmentation}
    
    Optical flow is often exploited for its ability in relating scene changes 
    along time-axis: He~\etal~\cite{yang_cvpr17} aggregates information from 
    multiple views using optical flow to perform semantic segmentation for 
    a single frame, while Zhu~\etal~\cite{zhu17dff} uses optical flow to 
    propagate image features from keyframe images to nearby frames, speeding
    up the segmentation process. Several methods exploit motion information
    to segment images into foreground objects from a moving background, 
    such as SegFlow~\cite{SegFlow_ICCV17} or FusionSeg~\cite{fusionseg}. However,
    they solely rely on the motion properties of objects and do not take the objects' 
    identities into account.
   
    In this paper, we leverage the notion of segmentation into a more semantic meaning of a scene,
    i.e. we do not limit the segmentation to just foreground/background~\cite{SegFlow_ICCV17}
    or coarse general classes (things, planes, stuff)~\cite{Sevilla-Lara_2016_CVPR}, but rather
    adhere to the current segmentation problems in the literature, which on average, consist of
    10-20 classes \cite{Cordts2016, virtualKITTI, PascalVOC}.
      
    As image motion is the projection of 3D object motions, depth discontinuities
    correspond to motion boundaries, making optical flow an indication of scene depth.
    At the same time, surface normals represent the changes of depth and the alignment
    of object surfaces. Such information is particularly useful for semantic segmentation, 
    as objects can be recognized not only from their appearances, but also from their shape and 
    geometric characteristics. Thus, similar to methods that recognize objects from depth images, 
    by associating each object with their motion type, or surface normals, 
    it is feasible to recognize them from surrounding regions using geometry information 
    signified from these modalities.
   

    \mysubsubsection{Three for Surface Normals}
    
    Similar to the case of optical flow, semantic segmentation provides object boundary 
    information, which in many cases corresponds to depth disruptions, thus helps to enhance
    object boundary, and local coherence in surface normal prediction. Ladicky~\etal~\cite{Ladicky2014} employs segment cues to enforce smooth surface
    normal results estimated with contextual information.
    
    Optical flow represents motion information, yet also signifies geometry  
    structure of a scene. As surface normals are identified by the rate of change in the location 
    of objects, they are highly correlated. Thus, optical flow can provide useful cues to
    enhance surface normals, in terms of objects' inner structure as well as their boundaries.
    

\section{Method}
\label{sec:method}
 
    To study the relationship between the three modalities, we follow a refinement strategy.
    That is based on the work by Jafari~\etal~ who uses a similar network to analyze the relationship between semantic segmentation and depth~\cite{Jafari2017}. 
    In this paper, we adapt their network architecture to train a joint refinement network designed for semantic segmentation, surface normals and optical flow.
    
\begin{figure}[t]
        \centering
        \includegraphics[width=.8\linewidth]{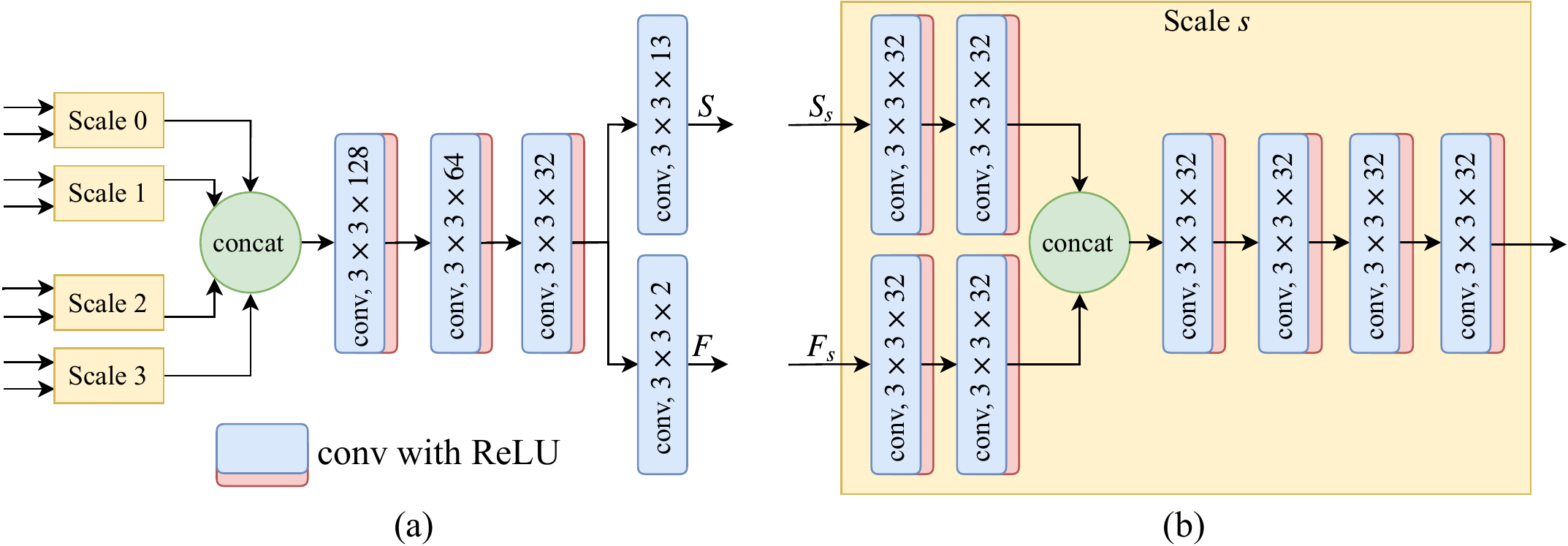}
        \vspace{2mm}
        \caption{Joint refinement network for two modalities with \emph{tight} feature coupling (\emph{left}), inspired by~\cite{Jafari2017}.
        The outputs of the modal-specific networks are integrated at different scales, using scale branch architecture (\emph{right}), and up-sampled before concatenation.
        }
        \label{fig:refine_module}
    \end{figure}

    An overview of the architecture is shown in Figure~\ref{fig:refine_module}a.
    As an example for joint refining optical flow and semantic segmentation, 
    the refinement network takes in a preliminarily predicted segmentation and optical 
    flow at different scales (Figure~\ref{fig:refine_module}b) and couples
    them in a joint optimization process.
    The input to the branch scale $s$ is composed of a segmentation image $S_s$ and a flow image
    $F_s$, both sub-sampled to  $\frac{1}{2^s}$ of the original size.
    The outputs of the scale branches are up-sampled, as the refinement network provides
    output at the original image size. We include a scale branch at the original size, and
    expand the network depth to increase its capacity to cope with different input modalities.
    When there are 3 modality inputs, the scale branch will have a third input, which is
    then concatenated to the other two.
    
    \begin{figure}[t]
        \centering
        \includegraphics[width=.8\linewidth]{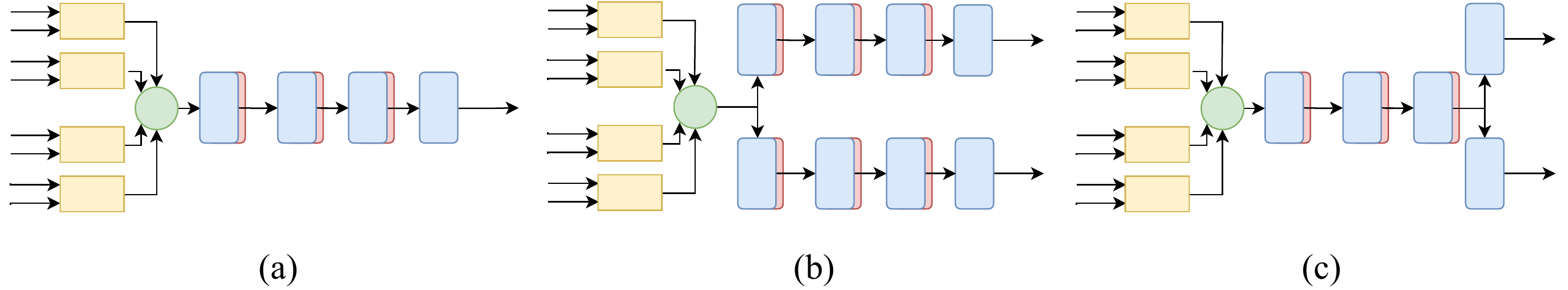}
        \vspace{2mm}
        \caption{Coupling levels of joint features in refinement:
        (a) \emph{zero} coupled, where joint features refine a single task;
        (b) \emph{loosely} coupled, where joint features branch to refine each task separately;
        (c) \emph{tightly} coupled, where joint features share a decoder to refine all tasks.
        }
        \label{fig:arch3s}
    \end{figure}    
    
    We also leverage the study of cross-modality influence performed in \cite{Jafari2017}, and
    the ability of the refinement network to learn a joint representation that benefits from both modalities.
    We keep  the multi-scale encoder part fixed (up to the \textit{concat} layer) and partially decouple the decoder.
    That allows the network to have different capacities in using the joint features learned from the encoder to refine
    different modalities. Specifically, we examine three different architectures that impose different coupling levels 
    of joint features as illustrated in Figure~\ref{fig:arch3s}. Namely, we study the ability of refinement
    when there is only one task required, hence \textit{zero}-coupling (Figure~\ref{fig:arch3s}a), when both 
    2 tasks are \textit{loosely} coupled (Figure~\ref{fig:arch3s}b), or \textit{tightly} coupled (Figure~\ref{fig:arch3s}c).


\section{Experiments}
\label{sec:experiments}

\subsection{Experimental Setup}
\mysubsubsection{Datasets}
    
    {}\hspace{5mm}\textbf{Nature}. We created a synthetic dataset covering nature scenes like gardens and parks. 
    The dataset features different vegetation types such as trees, bushes, flowering plants, and
    grass, being laid out in various terrain and landscape types. The models'
    textures and skies are used from real-world images to provide a realistic look of the scenes. 
    For each construction, we put up random paths to
    run the cameras around and capture the scene from different positions and angles.
    The images are rendered with the physics-based Blender
    Cycles engine\footnote{\url{https://www.blender.org/}}. To obtain annotations, 
    the rendering pipeline is modified to output RGB images, optical flow, surface normals, and objects' identities 
    (semantic labels). The dataset consist of 300 images from 10 scenes, each with 5 different lighting 
    conditions (clear, cloudy, overcast, sunset, and twilight), resulting in 15K images.
    
    \textbf{Virtual KITTI} \cite{virtualKITTI} (VKITTI) is a large-scale synthetic dataset following 
    the setting of KITTI dataset~\cite{Geiger2012} for autonomous driving problems. 
    
    Each image frame comes with pixel-wise annotation
    of ground truth instance-object segmentation, optical flow, and depth information. 
    To get surface normal ground truth, we convert ground truth depth images
    using the method described in~\cite{Barron2013}. As the depth images are produced by
    a simulation renderer, they are free from noise and uncertainties, producing less artifacts in the conversion results.

\mysubsubsection{Baselines \& Evaluation metrics}
Each of the three modalities have their own baseline and evaluation metric.

    \textbf{Optical Flow.} We use FlowNetC~\cite{Dosovitskiy2016} as our baseline
    because of its balance in speed and accuracy~\cite{fgfa_iccv17, SegFlow_ICCV17}
    (and is therefore preferable over the less accurate FlowNetS or the more expensive
    FlowNet2~\cite{Ilg2016}). We fine-tune the network for each dataset and report the results
    as baseline. Performance is evaluated by the average endpoint error, the lower the better.

    \textbf{Semantic Segmentation.} We use the ResNet-101 architecture~\cite{resnet} as the baseline, 
    and follow the practice of~\cite{SegFlow_ICCV17, Zhao2017} to add corresponding decoder layers 
    so that the network can output full-resolution image.
    Performance is measured by mean intersection-over-union, shown in percentage, the higher the better.
    
    \textbf{Surface Normals.} We follow the MarrRevisited~\cite{Bansal2016} architecture to run
    and report their results on our datasets as baseline. Evaluation is based on the angular 
    difference between predicted normals and ground truth~\cite{Fouhey2013}.
    The 3 error measurements \textit{mean},
    \textit{median}, \textit{rmse} show the difference (degrees) between predicted and 
    ground truth normal vectors, thus lower is better. The 3 measurements \textit{11.25}, 
    \textit{22.5}, \textit{30} count the number of pixels within the indicated angle thresholds
    (in degrees); the results are shown in percentage, and higher is better. 
   
\subsection{Baseline \& Oracle Experiments}

    The first set of experiments is to test the hypothesis that the modalities have a positive impact 
    on each other and to establish the baselines. For each experiment, we train the aforementioned 
    baseline networks, and have the output results passed into the refinement architecture (described
    in Section~\ref{sec:method}) together with ground truths of other modalities.
    
    \mysubsubsection{Optical Flow}
   
    Figure~\ref{tab:flow_gt} shows the baseline and refinement results for optical flow, 
    using ground truth segmentation and surface normals. 
    In general, both modalities help to improve optical flow. The refined results in
    Figure~\ref{fig:flow_gt} appear more crispy, especially along the objects' boundaries.
   
    \begin{figure*}	
    	\centering
    	
    	\captionsetup[subfigure]{labelformat=simple, labelsep=colon}
        \begin{subfigure}[t]{\linewidth}
           \renewcommand\thesubfigure{\alph{subfigure}}
           \captionsetup[subfigure]{labelformat=parens, labelsep=space}
            \begin{subfigure}[b]{\linewidth}
                \centering
                \setlength\doublerulesep{1pt}
                \resizebox{.5\columnwidth}{!}{%
                \begin{tabular}{c|c|c|c}
                    Dataset & FlowNetC~\cite{Dosovitskiy2016} & with GT seg & 
                        with GT norm \\ 
                    \hline
                    VKITTI & 2.68 & 2.37 & \textbf{2.36} \\ 
                    \hhline{====}
                    Nature & 16.19 & 14.09 & \textbf{13.92} \\ 
                \end{tabular}
                }
                \caption{Average EPE of optical flow baseline and oracle refinement, lower is better.}
  
                \label{tab:flow_gt}
            \end{subfigure}
        	\begin{subfigure}[b]{\linewidth}
                \centering
                \includegraphics[width=.7\linewidth]{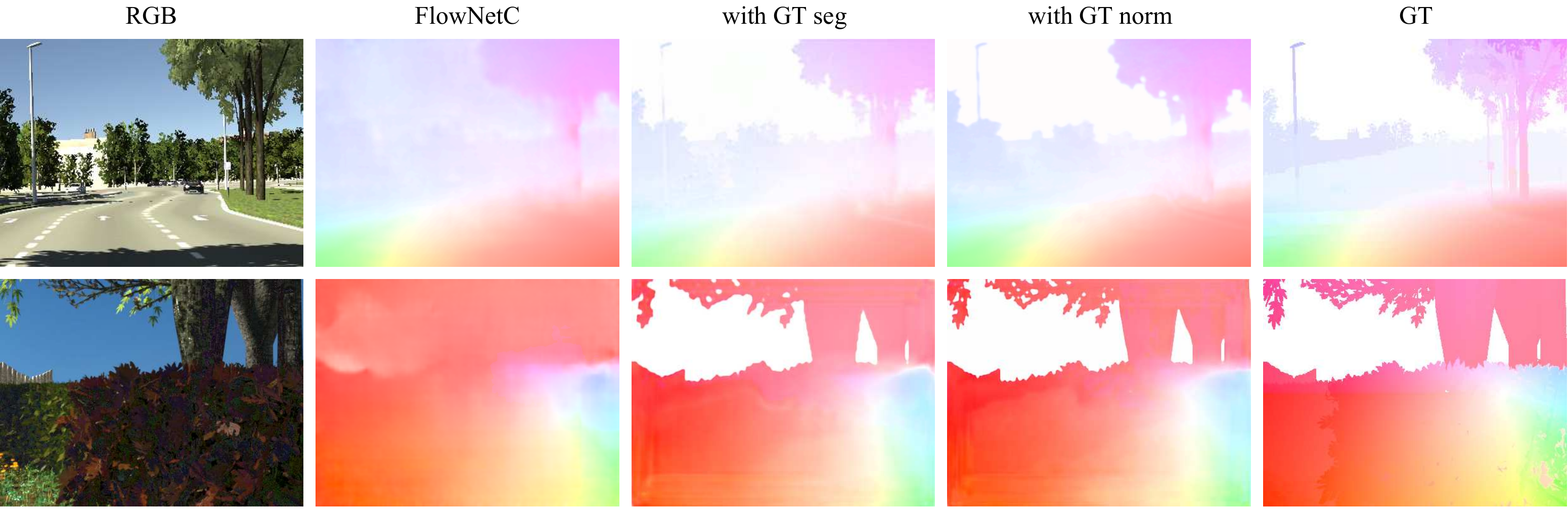}
                \caption{Qualitative examples of optical flow on VKITTI (\emph{top}) and Nature (\emph{bottom})}.
  
                \label{fig:flow_gt}
        	\end{subfigure}
    	
            \renewcommand\thesubfigure{Figure \arabic{figure}}
            \vspace{-3mm}
        	\caption{Results of optical flow oracle refinement: adding ground truth
        	segmentation and surface normal improves the crispness of objects' boundaries
        	and outperforms all the baselines.}
            \label{subfig:flow_gt}
        \end{subfigure}
        \addtocounter{figure}{+1}
        \addtocounter{subfigure}{-3}
            
        \vspace{+1mm}
    	\captionsetup[subfigure]{labelformat=simple, labelsep=colon}
        \begin{subfigure}[t]{\linewidth}
           \renewcommand\thesubfigure{\alph{subfigure}}
           \captionsetup[subfigure]{labelformat=parens, labelsep=space}
           \begin{subfigure}[b]{\linewidth}
                \centering
                \setlength\doublerulesep{1pt}
                \resizebox{.5\columnwidth}{!}{%
                \begin{tabular}{c|c|c|c}
                    Dataset & ResNet~\cite{SegFlow_ICCV17} & with GT flow & with GT norm  \\
                    \hline
                    VKITTI & 44.11 & 46.90 & \textbf{50.0} \\
                    \hhline{====}
                    Nature & 37.88 & 38.4 & \textbf{41.6}
                \end{tabular}
                }
                \caption{
                Mean IOU (in \%) of segmentation baseline and oracle refinement, 
                higher is better.
                }
                \label{tab:seg_gt}
            \end{subfigure}
     
        	\begin{subfigure}[b]{\linewidth}
                \centering
                \includegraphics[width=.7\linewidth]{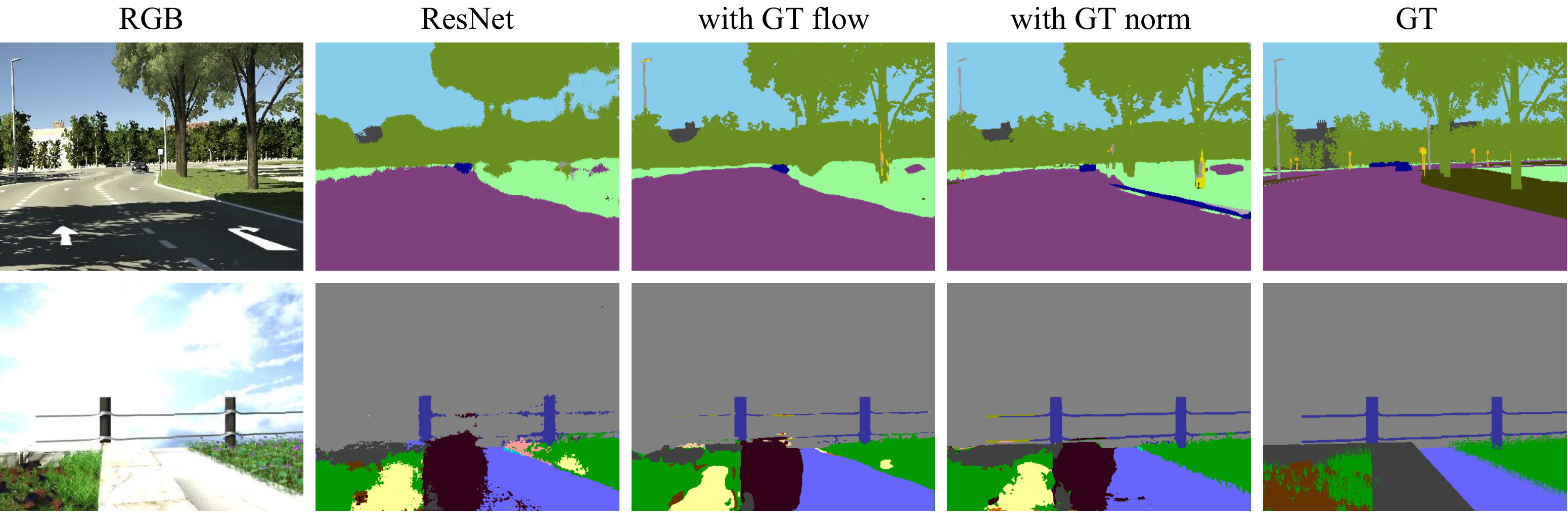}
                \caption{
                Qualitative examples of segmentation on VKITTI (\emph{top}) and Nature \emph{bottom}.
                }
                \label{fig:seg_gt}
        	\end{subfigure}
            \renewcommand\thesubfigure{Figure \arabic{figure}}
            \vspace{-4mm}
            \caption{Results of segmentation oracle refinement: adding ground truth flow or surface normals captures more fine-detailed segmentations (\emph{e.g.} tree leaves and fence wires) and improves performance over the baseline, since the shape of the objects obtained is informative.}
            \label{subfig:seg_gt}
            \addtocounter{figure}{+1}
            \addtocounter{subfigure}{-3}
        \end{subfigure}
        
    	\captionsetup[subfigure]{labelformat=simple, labelsep=colon}
        \begin{subfigure}[b]{\linewidth}
           \captionsetup[subfigure]{labelformat=parens, labelsep=space}
            \begin{subfigure}[b]{\linewidth}
               
                \centering
                \setlength\doublerulesep{1pt}
                \resizebox{.7\columnwidth}{!}{%
                \begin{tabular}{c|c|c|c|c|c|c|c}
                        \multicolumn{1}{c}{} & & mean ($\downarrow$)& median ($\downarrow$) & rmse ($\downarrow$) & 11.25 ($\uparrow$) & 22.5 ($\uparrow$) & 30 ($\uparrow$) \\
                        \hline
                        \parbox[t]{2mm}{\multirow{3}{*}{\rotatebox[origin=c]{90}{\small \textbf{VKITTI}}}} & MarrRevisited~\cite{Bansal2016} & 48.13 & 48.34 & 57.44 & 17.39 & 19.81 & 27.44 \\
                        & with GT flow & 11.26 & 3.18 &	17.29 &	\textbf{62.17} & \textbf{76.27} & 87.14  \\
                        & with GT seg &	\textbf{11.16} & \textbf{3.17} & \textbf{16.78} & 60.70 & 75.67 & \textbf{88.43} \\
                    \hhline{========}
                        \parbox[t]{2mm}{\multirow{3}{*}{\rotatebox[origin=c]{90}{\textbf{Nature}}}} & MarrRevisited~\cite{Bansal2016} & 40.25 & 46.68 & 50.25 & 29.44 & 30.68 & 34.42 \\
                        & with GT flow & \textbf{9.62} & \textbf{8.01} & \textbf{13.20} & \textbf{61.78} & \textbf{89.74} &	\textbf{97.53} \\
                        & with GT seg & 10.39 &	8.75 & 13.48 & 59.65 & 89.46 & 97.30 \\
                \end{tabular}
                }
                \caption{
                Quantitative results of surface normals baseline and oracle refinement. 
                }
                \label{tab:norm_gt}
            \end{subfigure}
     
        	\begin{subfigure}[b]{\linewidth}
                	 
                \centering
                \includegraphics[width=.7\linewidth]{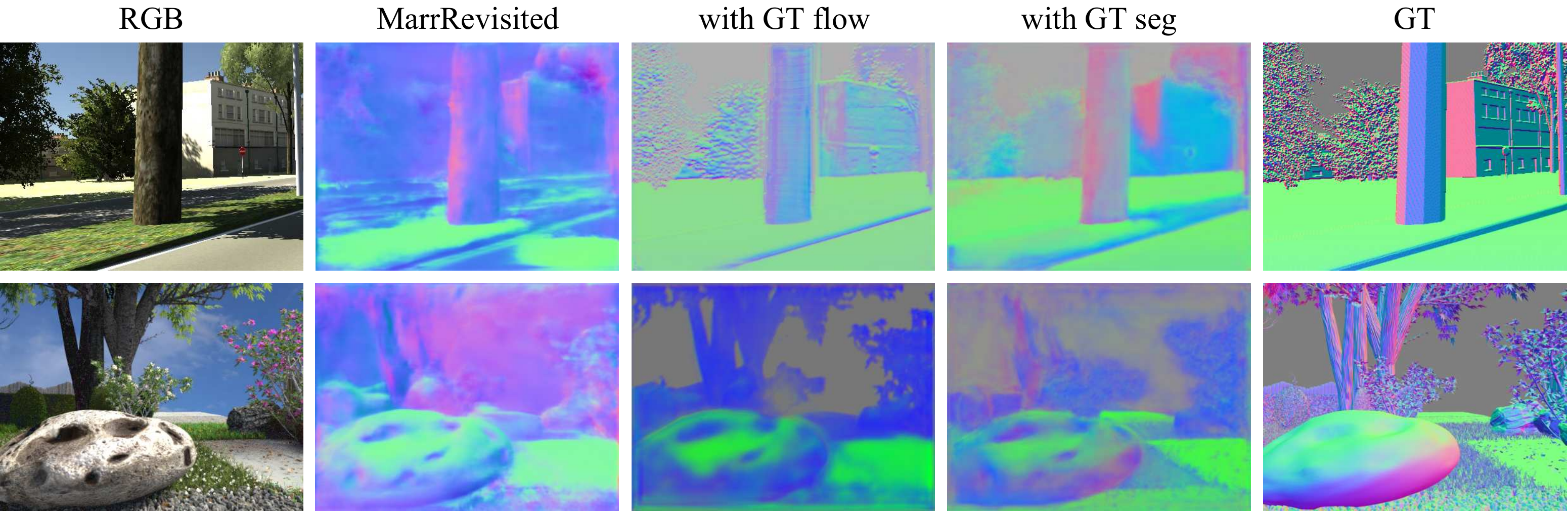}
                \caption{
                Qualitative examples of surface normal on VKITTI (\emph{top}) and Nature (\emph{bottom}).
                }
                \label{fig:norm_gt}
        	\end{subfigure}
            \renewcommand\thesubfigure{Figure \arabic{figure}}
            \vspace{-4mm}
            \caption{Results of surface normals oracle refinement: adding ground truth flow and segmentation removes the blur in the baseline and improve the results, indicating the interest of combining modalities.}
            \label{subfig:norm_gt}
        \end{subfigure}
    \end{figure*}

    \mysubsubsection{Semantic Segmentation}

    As shown in Figure~\ref{tab:seg_gt}, optical flow and surface normals also improve 
    semantic segmentation over the baseline, since the outline of the objects obtained from these
    modalities are more informative. The geometry information provided by flow and normals helps to
    retain details in semantic segmentation; e.g. the lamppost, tree branches, and fence 
    wires are well retained in Figure~\ref{fig:seg_gt}. However, as the refinement module does not
    have access to the original image (raw RGB), geometry information alone cannot help much in 
    correcting semantic errors that are present in the input (yellow regions in the second row).
 
    \mysubsubsection{Surface Normals}

    As shown in Figure~\ref{tab:norm_gt}, using optical flow and semantic segmentation helps
    to imrpove surface normal prediction significantly. The refinement using flow seems to be better
    than using segmentation for the VKITTI case, whereas it is the other way around for the Nature case.
    This can be explained as estimating surface normals requires a network to understand the 
    geometry information of the scene, which is easier for optical flow in Nature as all of the objects 
    are static and thus  optical flow field depends solely on the camera ego-motion, while it is not the 
    case for VKITTI, segmentation has more advantage as objects' shapes are more uniform (e.g. houses, cars,
    roads' surface) than those in Nature (e.g. bushes, rocks, grass).
    The refined result using oracle flow produces sharper details, while the ones with oracle segmentation 
    are more accurate (Figure~\ref{fig:norm_gt}).

    \vspace{3pt}
    \noindent\textbf{To conclude}, different modalities, when being used in their most accurate form (GT),
    provide complementary cues to each other, thus improving the performance of other modalities:
    segmentation provides flows and normals with objects' identities and boundaries;
    optical flow provides segmentation and normals motion and geometry information;
    surface normals provide segmentation and flow with geometry and objects boundaries.
    In the following experiments, we examine the usefulness of different modalities when 
    they are not perfect (given by an approximation scheme), and the interaction of more than one modalities.

\subsection{Cross-Modality Influence}

\mysubsubsection{Refinement Coupling}
 
    In this experiment, we consider joint learning of semantic segmentation and optical flow on the VKITTI dataset~\cite{virtualKITTI}.
    We examine the different coupling levels during refinement: \textit{zero} coupling, \textit{loose} coupling, and
    \textit{tight} coupling (see Figure~\ref{fig:arch3s}). 
    We also expand the \emph{tight} coupling, into an end-to-end learning pipeline (denoted by \textit{tight+}), 
    where the segmentation and optical flow networks are fine-tuned together with the refinement module.
    
    \begin{table*}[t]
        \centering
        \setlength\doublerulesep{1pt}
        \resizebox{.7\columnwidth}{!}{%
            \begin{tabular}{c|c|c||c|c|c|c}
                \hline
                \multirow{2}{*}{Target} & \multirow{2}{*}{Baseline} & GT & 
                    \multicolumn{4}{c}{Predicted} \\
                    \cline{3-7}
                & & \textit{zero} & \textit{zero} & \textit{loose} & \textit{tight} & \textit{tight}+  \\
                    \hline
                Semantic segmentation ($\uparrow$)  & 44.11  & {46.9} & \textbf{44.78} & 41.2 & 41.1 & 43.9 \\            
                    \hhline{=======}
                Optical flow ($\downarrow$)   & 2.68 & {2.37} & \textbf{2.40} & 2.43 & 2.41 & 2.42  \\
            \end{tabular}
        }
        \vspace{+2mm}
        \caption{
        Different levels of coupling of segmentation and optical flow on the VKITTI dataset.
        Performance is measured in Mean IOU ($\uparrow$) and Average EPE ($\downarrow$) respectively.
        Refinement can improve over the baseline, and \textit{zero} coupling, when both modalities refine a single task, works best for both semantic segmentation and optical flow.
        }
        \label{tab:refine_compare}
    \end{table*}
    
    The results are shown in Table~\ref{tab:refine_compare}.
    From the results, we observe that using ground-truth or predicted segmentation to refine optical flow, the performance always improves. The difference between ground-truth and predicted segmentation is small compared to the difference between the baseline and the refined models.
    However, for segmentation, refining based on flow is only beneficial when the \emph{zero} coupling is used. 
    Likely, this is because the predicted flow does not contain accurate semantic cues to improve segmentation.
    Based on this experiment we use the \emph{zero} coupling for the remaining experiments.

\mysubsubsection{Flow from Segmentation and Normals}

    The refinement results for optical flow using predicted modalities are shown in 
    Figure~\ref{tab:flow_pr}. Because of inaccuracies in predicted normals, the refined results
    do not improve as much as with predicted segmentation, or even hurt in case of Nature dataset.
    In general, the two modalities help optical flow to obtain better delineation. 
    Figure~\ref{fig:flow_pr} shows an occlusion case where part of the car is occluded by a traffic sign, 
    FlowNetC recognizes it but fails to obtain the correct shape of the occlusion, which can be recovered 
    with surface normal information and improved using segmentation and surface normals.

    \begin{figure*}	
    	\centering
    	\captionsetup[subfigure]{labelformat=simple, labelsep=colon}
        \begin{subfigure}[t]{\linewidth}
           \renewcommand\thesubfigure{\alph{subfigure}}
           \captionsetup[subfigure]{labelformat=parens, labelsep=space}
            \begin{subtable}[b]{\linewidth}
                \centering
                \setlength\doublerulesep{1pt}
                \resizebox{.7\columnwidth}{!}{%
                \begin{tabular}{c|c|c|c|c}
                    \hline
                        Method & FlowNetC~\cite{Dosovitskiy2016} & with PR seg & with PR norm & with PR seg+norm \\
                        \hline
                        VKITTI & 2.68 & 2.40 & 2.50 & \textbf{2.39}  \\
                        \hhline{=====}
                        Nature  & 16.19 & \textbf{14.16} & 16.62 & 14.21  \\
                        \hline
                \end{tabular}
                }
                \caption{Average EPE of optical flow based on prediction, lower is better.}
                \label{tab:flow_pr}
            \end{subtable}
        	\begin{subfigure}[b]{\linewidth}
        		\centering
        		\includegraphics[width=.8\textwidth]{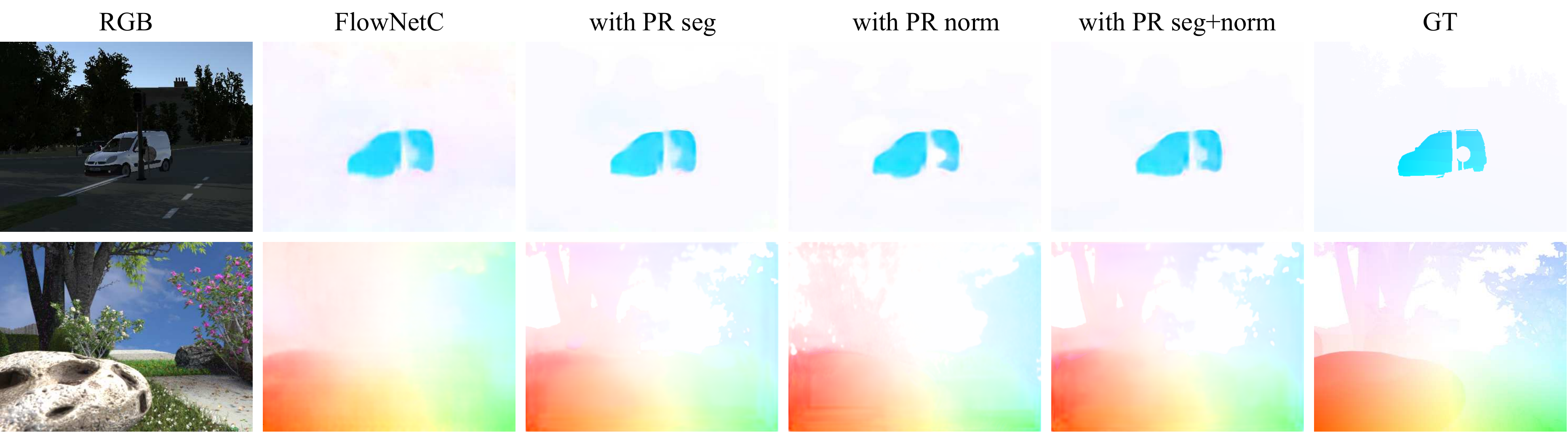}
        	    \caption{Qualitative examples on VKITTI (\textit{top}) and Nature (\textit{bottom}).}
        		\label{fig:flow_pr}		
        	\end{subfigure}
    	
            \renewcommand\thesubfigure{Figure \arabic{figure}}
            \vspace{-1mm}
        	\caption{Results of optical flow baseline and refinement based on prediction.}
            \label{subfig:flow_pr}
            \vspace{+3mm}
        \end{subfigure}
        \addtocounter{figure}{+1}
        \addtocounter{subfigure}{-3}
            
    	\captionsetup[subfigure]{labelformat=simple, labelsep=colon}
        \begin{subfigure}[t]{\linewidth}
           \renewcommand\thesubfigure{\alph{subfigure}}
           \captionsetup[subfigure]{labelformat=parens, labelsep=space}
           \begin{subfigure}[b]{\linewidth}
                \centering
                \setlength\doublerulesep{1pt}
                \resizebox{.7\columnwidth}{!}{%
                \begin{tabular}{c|c|c|c|c}
                    \hline
                        Method & ResNet~\cite{SegFlow_ICCV17} & with PR flow & with PR norm & with PR flow+norm \\
                        \hline
                        VKITTI & 44.11 & 44.78  & 45.36 & \textbf{47.55}  \\
                        \hhline{=====}
                        Nature  & 37.88 & 37.57  & \textbf{38.83} & 38.00  \\
                        \hline
                \end{tabular}
                }
                \caption{Mean IOU (in \%) of segmentation based on prediction, higher is better.}
                \label{tab:seg_pr}
            \end{subfigure}
     
        	\begin{subfigure}[b]{\linewidth}
        		\centering
        		\includegraphics[width=.8\textwidth]{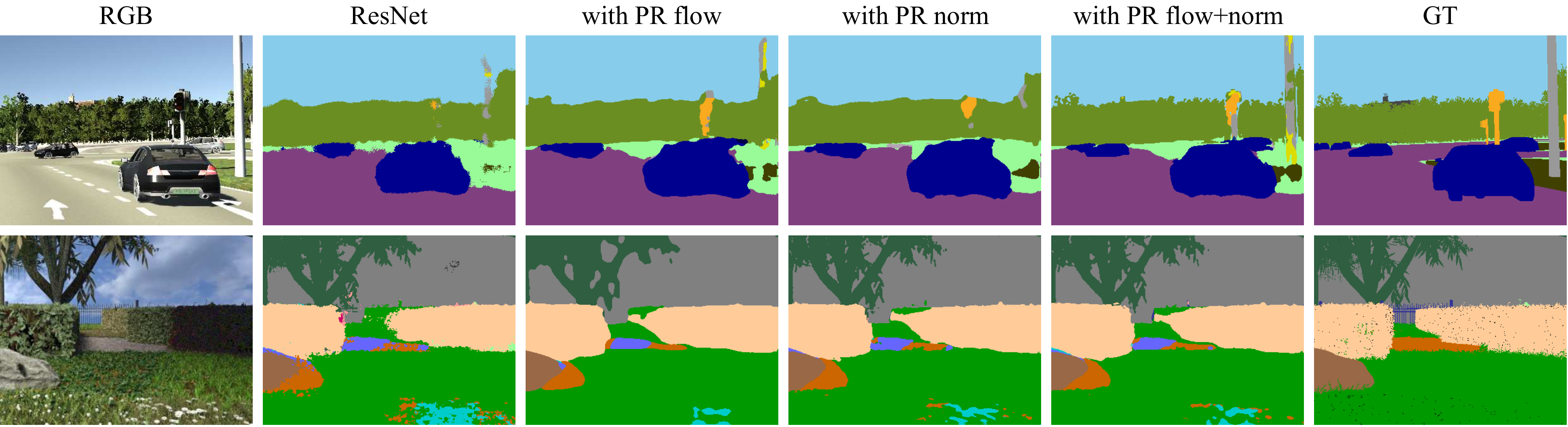}
        	    \caption{Qualitative examples on VKITTI (\textit{top}) and Nature (\textit{bottom}).}
        		\label{fig:seg_pr}
        	\end{subfigure}
            \renewcommand\thesubfigure{Figure \arabic{figure}}
            \vspace{-4mm}
            \caption{Results of segmentation baseline and refinement based on prediction.}
            \label{subfig:seg_pr}
            \vspace{+2mm}
            \addtocounter{figure}{+1}
            \addtocounter{subfigure}{-3}
        \end{subfigure}
        
    	\captionsetup[subfigure]{labelformat=simple, labelsep=colon}
        \begin{subfigure}[b]{\linewidth}
           \captionsetup[subfigure]{labelformat=parens, labelsep=space}
            \begin{subfigure}[b]{\linewidth}
                \centering
                \setlength\doublerulesep{1pt}
                \resizebox{.75\columnwidth}{!}{%
                \begin{tabular}{c|c|c|c|c|c|c|c }
                        & Method 
                         & mean ($\downarrow$)& median ($\downarrow$) & rmse ($\downarrow$) & 11.25 ($\uparrow$) & 22.5 ($\uparrow$) & 30 ($\uparrow$) \\
                        \hline
                        \parbox[t]{2mm}{\multirow{4}{*}{\rotatebox[origin=c]{90}{\small \textbf{VKITTI}}}} & MarrRevisited~\cite{Bansal2016} & 48.13 & 48.34 & 57.44 & 17.39 & 19.81 & 27.44 \\
                        & with PR flow & 12.32 & 4.91 &	18.02 &	58.43 & 73.46 &	84.08 \\
                        & with PR seg & \textbf{11.44} & \textbf{2.82} & \textbf{17.24} & \textbf{60.30} & 74.02 & 86.58 \\
                        & with PR flow+seg &	11.58 & 3.45 & 17.28 & 60.04 & \textbf{74.63} & \textbf{86.59} \\
                    \hhline{========}
                        \parbox[t]{2mm}{\multirow{3}{*}{\rotatebox[origin=c]{90}{\textbf{Nature}}}} & MarrRevisited~\cite{Bansal2016} & 40.25 & 46.68 & 50.25 & 29.44 & 30.68 & 34.42 \\
                        & with PR flow & 11.39 & 9.80 & 14.38 & 55.36 & 86.88  & 96.66	 \\
                        & with PR seg & \textbf{9.09} & \textbf{7.08} & \textbf{12.56} & \textbf{65.26} & \textbf{91.11} & \textbf{97.70} \\
                        & with PR flow+seg & 9.22 & 7.27 & 12.71 & 64.39 & 90.94 & 97.59 \\
                        \hline
                \end{tabular}
                }
                \caption{Quantitative results of surface normal refinement based on prediction.
               }
                \label{tab:norm_pr}

            \end{subfigure}
     
        	\begin{subfigure}[b]{\linewidth}
        		\centering
        		\includegraphics[width=.8\textwidth]{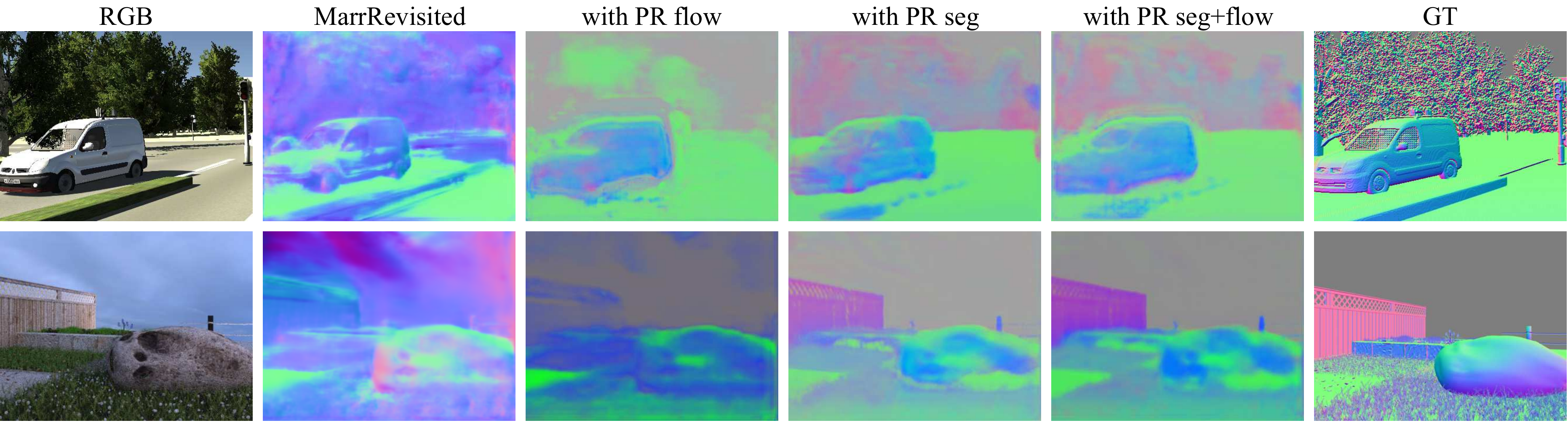}
        		\caption{Qualitative examples on VKITTI (\textit{top}) and Nature (\textit{bottom}).}\label{fig:norm_pr}
        	\end{subfigure}
            \renewcommand\thesubfigure{Figure \arabic{figure}}
            \vspace{-4mm}
            \caption{Results of surface normals baseline and refinement based on prediction.}
            \label{subfig:norm_pr}
            \vspace{+2mm}
        \end{subfigure}
    \end{figure*}

\mysubsubsection{Segmentation from Flow and Normals}

    Predicted optical flow and surface normals contain different inaccuracies. Thus, when used to refine the
    semantic segmentation, they confuse the semantic cues, and to some extent, have negative impact
    on the preliminary segmentation results. This explains the decreasing results and slight improvements in
    Figure~\ref{tab:seg_pr}. Visual inspection on Figure~\ref{fig:seg_pr} shows
    that refinement of flow and surface normals make the boundaries smoother, reducing
    the effect of incorrect areas. In combination, they capture more details and help 
    to produce better segmentation.
     
\mysubsubsection{Normals from Flow and Segmentation}

    Surface normal refinement results are provided in Figure~\ref{tab:norm_pr} and illustrated in 
    Figure~\ref{fig:norm_pr}. The confusion in the sky and tree regions of the baseline estimation is
    removed when refined with different modalities. 
    Information of objects' categories and boundaries provided by semantic segmentation helps retaining fine details in the results (e.g. pavement in VKITTI, fences in Nature).
    However, the inaccuracies of predicted flow
    leave some artifacts and makes the results less accurate (e.g. the sky, tree and fences regions).
    Surprisingly, refinement using predicted segmentation outperform refinement with ground truth in Figure~\ref{tab:norm_gt}. 
    This could be explained by the idea of knowledge distillation~\cite{Hinton2015}, the predicted segmentation with smoothed softmax scores, describes the knowledge of the network better than the hard one-hot encoding of the ground-truth. This allows the refinement module to learn better.

\section{Conclusion and Future Work}
\label{sec:conclusion}

We have analyzed the combination of three important modalities in computer vision, namely optical flow, semantic segmentation,
and surface normals, and their impact on each other. Because each modality contains different type of
information, in combination, they provide complementary cues to enhance each other. 
We approached the problem at a modular level where the inputs are kept fixed at the
preliminary estimation. Future work will include end-to-end training of modalities to exploit raw image features.

\vspace{3pt}
\noindent\textbf{Acknowledgements}: This project was funded by the EU Horizon 2020 program No.
688007 (TrimBot2020). We would like to thank Leo Dorst for his helpful discussions and advice.

\bibliography{UvA-bmvc18}
\end{document}